\documentclass[runningheads]{llncs}

\usepackage[T1]{fontenc}
\usepackage{graphicx}

\usepackage{amsmath}
\newcommand{\methodeName}{Leveraging Inpainting for Local Interpretability }
\makeatletter
\newcommand\footnoteref[1]{\protected@xdef\@thefnmark{\ref{#1}}\@footnotemark}
\makeatother

\begin{document}

\title{Inpainting Insights: Elevating Visual XAI with Photorealistic Perturbations}

\titlerunning{Inpainting Insights}

\author{Josef Lindl\inst{1}\orcidID{0009-0007-6656-0632} \and
Mariana Chaves\inst{2}\orcidID{0009-0001-8086-9034} \and
Damien Garreau\inst{1}\orcidID{0000-0002-7855-2847}
}

\authorrunning{J. Lindl et al.}

\institute{
    Center for Artificial Intelligence and Data Science (CAIDAS), Julius-Maximilans-Universit\"at W\"urzburg, Sanderring 2, 97070 W\"urzburg, Germany \\
    \email{lindl.josef@googlemail.de} \\
    \email{damien.garreau@uni-wuerzburg.de}\\
    \and
    University of Groningen, Broerstraat 5, 9712 CP Groningen, The Netherlands \\
    \email{m.e.chaves.espinoza@rug.nl}
}

\maketitle

\begin{abstract}
The increasing complexity of state-of-the-art machine learning models has made their behavior progressively harder to interpret, 
spurring rapid advancements in the field of eXplainable Artificial Intelligence (XAI).
Among many methods proposed, perturbation-based approaches play a major role. 
By systematically altering (perturbing) input features, these approaches measure the impact on the model's predictions.
For image data, traditional perturbation techniques, often involve replacing pixel values \emph{e.g.}, with a pre-defined color.
However, such approaches, but also more refined deterministic techniques, 
generate unrealistic out-of-distribution samples and often leave visible artifacts,
which can mislead the model and compromise explanation quality. 
In this work, we adjust LIME, a widely used perturbation-based method,
to demonstrate how generative inpainting can improve perturbation-based explanations for images.
We achieve photorealistic perturbed samples that align better with the original data distribution and enhance explanation quality.
\end{abstract}

\section{Introduction} \label{sec:introduction}

As Artificial Intelligence (AI) systems increasingly influence critical decision-making,
the need for transparency and accountability has become paramount \cite{PAPAGIANNIDIS2025101885,Radanliev31122025,10.3389/fhumd.2024.1421273}.
For example computer vision models can achieve high accuracy by exploiting unintended dataset artifacts rather than learning meaningful visual features
\cite{Lapuschkin2019}. The research field of eXplainable AI (XAI) explores methods, to make complex machine learning models understandable to humans,
fostering trust and ethical deployment. 

Perturbation-based XAI methods, such as LIME (Local Interpretable Model-agnostic Explanations) 
\cite{lime}, SHAP (SHapley Additive exPlanations) \cite{NIPS2017_7062}, 
and RISE (Randomized Input Sampling for Explanation) \cite{DBLP:conf/bmvc/PetsiukDS18},
have become foundational tools for interpreting image classifiers \cite{s25134166}. 
These approaches generate perturbed samples by systematically removing or occluding parts of the input image, 
replacing them with a constant color (\emph{e.g.}, gray, or mean pixel value),
blurring, or zeroing out pixels \cite{lime,10.1007/978-3-319-10590-1_53,DBLP:conf/bmvc/PetsiukDS18}.

As an example, LIME for images segments the input image into superpixels 
and creates perturbed samples through occlusion, replacing them with the mean color of the superpixel by default.
However, LIME perturbation sampling can be adopted for different replacement methods, 
as we demonstrate in Section~\ref{sec:deterministic-occlusion} by 8 deterministic occlusion strategies. 
Still, we can show, that there are cases where all of these approaches fail.
More generally, despite being widely used, deterministic perturbation methods tend 
to introduce unnatural artifacts and generate out-of-distribution samples. This can lead to
unpredictable model behavior and ambiguous attributions \cite{aghakishiyeva2025photorealistic,10.1007/978-3-031-63787-2_12}.

A shared intuition behind replacement approaches is that occlusion should create plausible content in place of the
removed region. Agarwal and Nguyen \cite{agarwal2020explaining}
demonstrated that generative inpainting models can maintain the original data distribution while removing discriminative features.
Their adaption of LIME, LIME-G, uses inpainting with DeepFill v1 and v2 \cite{yu2018generative,yu2018free} to generate perturbed samples. 
However, more recent models, such as LaMa \cite{9707077}, outperform DeepFill in perceived inpainting quality.
Nevertheless, despite these advancements, no work has yet explored improving LIME-G with state-of-the-art inpainting models.

Our contributions are as follows:
Firstly, we introduce \textbf{\methodeName (LILI)},
a method that incorporates generative perturbation mask inpainting to explain images using LIME. 
Secondly, by leveraging overlapping superpixels (perturbation mask expansion),
we obscure residual artifacts and outlines for more effective occlusion.
We demonstrate that mask expansion helps mitigate the unintended reconstruction of occluded features
by adapting the perturbation process to the generative capabilities of modern inpainting models.
Finally, our evaluation reveals that LILI produces more realistic perturbations regarding perceived image quality,
favoring more consistent distributions of high-level features deep within the explained network.
Moreover, we show that LILI achieves improved explanation quality compared to LIME and LIME-G,
highlighting the effectiveness of using advanced generative inpainting models for perturbation-based explanations.

\begin{figure}[h!] 
\begin{center}
\includegraphics[clip=true,width=0.8\textwidth]{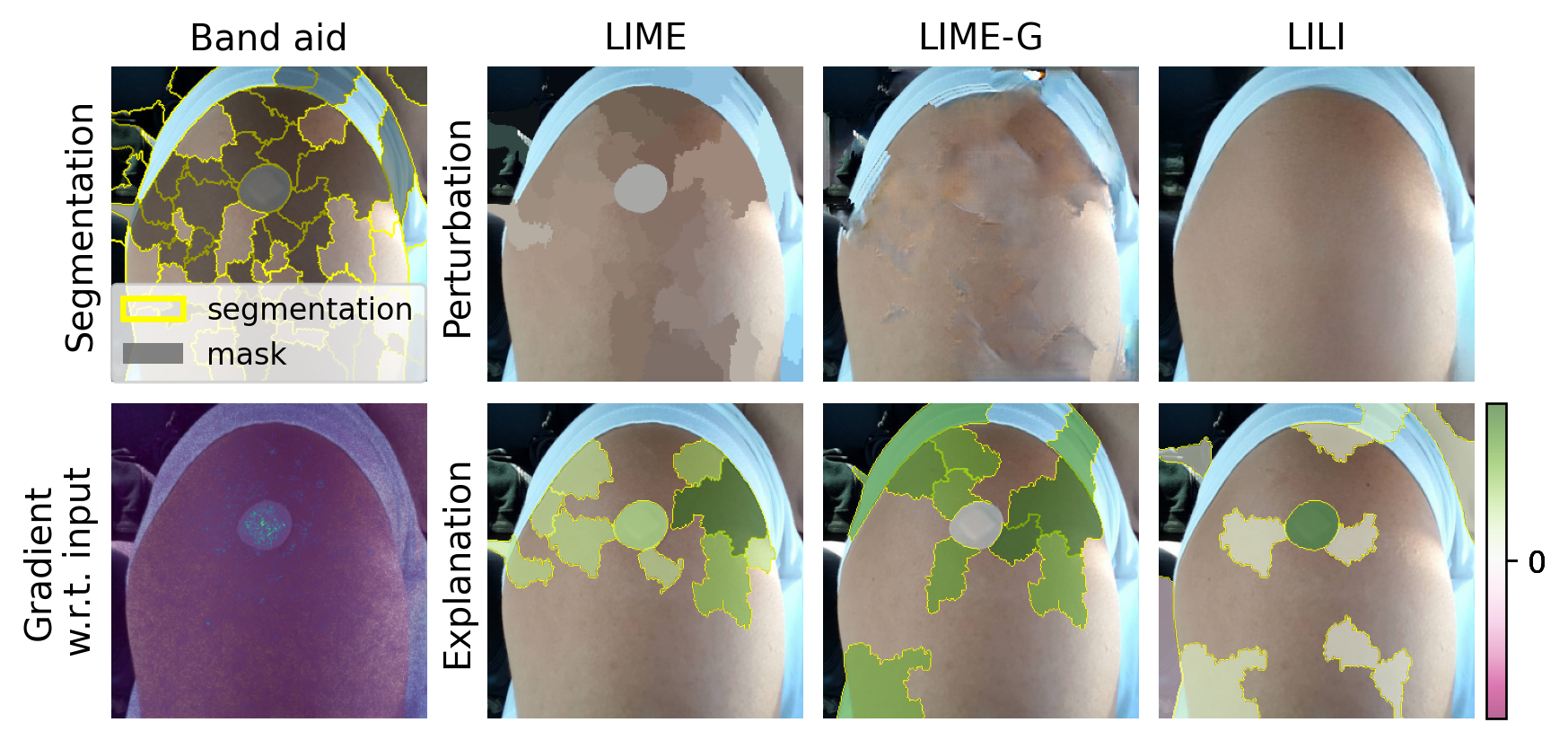}
\caption{
	\textbf{Perturbed samples and explanations for \textit{band-aid} from different methods:} 
    The first row shows the segmented \textit{original} image with an example of a perturbation mask (left image), 
    as well as the perturbed samples as given by \textit{LIME}, \textit{LIME-G} and \textit{LILI}.
    The second row shows the model's gradients with respect to the input, as well as the explanations given by different methods
    (here, the top ten superpixels by absolute weight are visualized: darker green indicates a stronger positive contribution). 
    Note that the gradient w.r.t. the input reveals that the band-aid superpixel seems to influence the prediction. 
    However, only LILI identifies the band-aid as most important.
}
\label{fig:explanation-compare-introduction}
\end{center}
\end{figure}

Fig.~\ref{fig:explanation-compare-introduction} shows an example of perturbed samples,
as well as explanations as given by LIME-G, and LILI. Note that only LILI identifies the band-aid as most important.
All the code for the method and the evaluation is open source and available on 
GitHub\footnote{https://github.com/jo01123/LILI}\fnmsep\footnote{\label{fn:github-chaves}https://github.com/m-chaves/LIME}.

\section{Background and Motivation} \label{sec:theoretical-background}

In this section, we will outline the foundations for the proposed method LILI. Firstly, we introduce
LIME in Section~\ref{sec:lime}. Secondly, in Section~\ref{sec:deterministic-occlusion}, we
discuss examples, where deterministic occlusion fails.
Section~\ref{sec:lama} introduces the LaMa inpainting model and discusses why it was 
selected for LILI. Finally, in Section~\ref{sec:related-work}, we present related work.

\subsection{LIME}\label{sec:lime}

Ribeiro et al.\ \cite{lime} introduced Local Interpretable Model-agnostic Explanations (LIME),
a method for explaining individual predictions (local) of any black-box model (model-agnostic).
The core idea is to approximate the complex model locally with an interpretable surrogate model, such as linear regression.
For this, LIME perturbs the input data by modifying interpretable features and observes the resulting changes in the model's output.
The surrogate model is then fitted to these changes to estimate feature importance. 
For image data, LIME defines interpretable features as superpixels, groups of pixels with similar properties.
Perturbed samples are created by randomly masking out some of these superpixels, 
allowing the surrogate model to approximate the behavior of the original model around the specific instance being explained.

Following the notation of Garreau and Mardaoui \cite{garreau2021doeslimereallyimages},
we can describe the core steps of LIME for images as follows. Let $f$ be the model and $\xi$ the sample to explain.
\begin{enumerate}
    \item \textit{Create interpretable features} by segmenting the image into $d$ disjunct superpixels 
    $J_1, \dots, J_d$ using a segmentation algorithm (\emph{e.g.}, quickshift \cite{10.1007/978-3-540-88693-8_52} by default).
    \item \textit{Create perturbed samples} $x_1, \dots, x_n$.
    For each $x_i$ draw $z_i \in \{0,1\}^d$ from independent Bernoulli trials.
    For all $z_{i,j} = 1$ the superpixel $J_j$ is masked for sample $x_i$ \emph{i.e.}, switched off.
    Replace therefore all pixel of masked superpixels with the superpixel's mean color (default) or a replacement color.
    An example of a segmentation, a perturbation mask and a perturbed sample is given in Fig.~\ref{fig:explanation-compare-introduction}.
    \item \textit{Compute weights $w_i$ from $x_i$s proximity to $\xi$} \emph{e.g.}, using an exponential kernel.
    Perturbed samples closer to the original (\emph{i.e.}, with less superpixels switched off) receive a higher weight. 
    \item \textit{Query the model} by making predictions $y_i = f(x_i)$.
    \item \textit{Fit the local surrogate model} to the~$y_i$s (target) on the presence or absence of superpixels ($z_i$s).
    By default, Ridge regression weighted by~$w_i$ is used. 
    The resulting coefficients of the surrogate model can be interpreted as the approximated contributions 
    of the corresponding superpixels to the prediction of the model $f$ on the original image~$\xi$.
    By mapping the pixels of the image to its superpixel weights, we receive the attribution map, 
    which can be visualized as heatmap.
\end{enumerate}

\subsection{Where Deterministic Occlusion Fails}\label{sec:deterministic-occlusion}

A natural idea is to improve superpixel replacement with a better, but simple method.
For instance, we tested eight deterministic approaches for LIME (including
the native LIME replacement by mean superpixel color or black):
(1) mean color of the superpixel, (2) mean image color, (3) mean color of neighboring superpixels,
(4) zeroing out (black) (5) median color of pixels at the image border (6) median color of superpixels at the image border
(7) median color of superpixel's neighbors mean, and (8) median color of the superpixel means.

\begin{figure}[h!]
\centering
\includegraphics[width=0.8\textwidth]{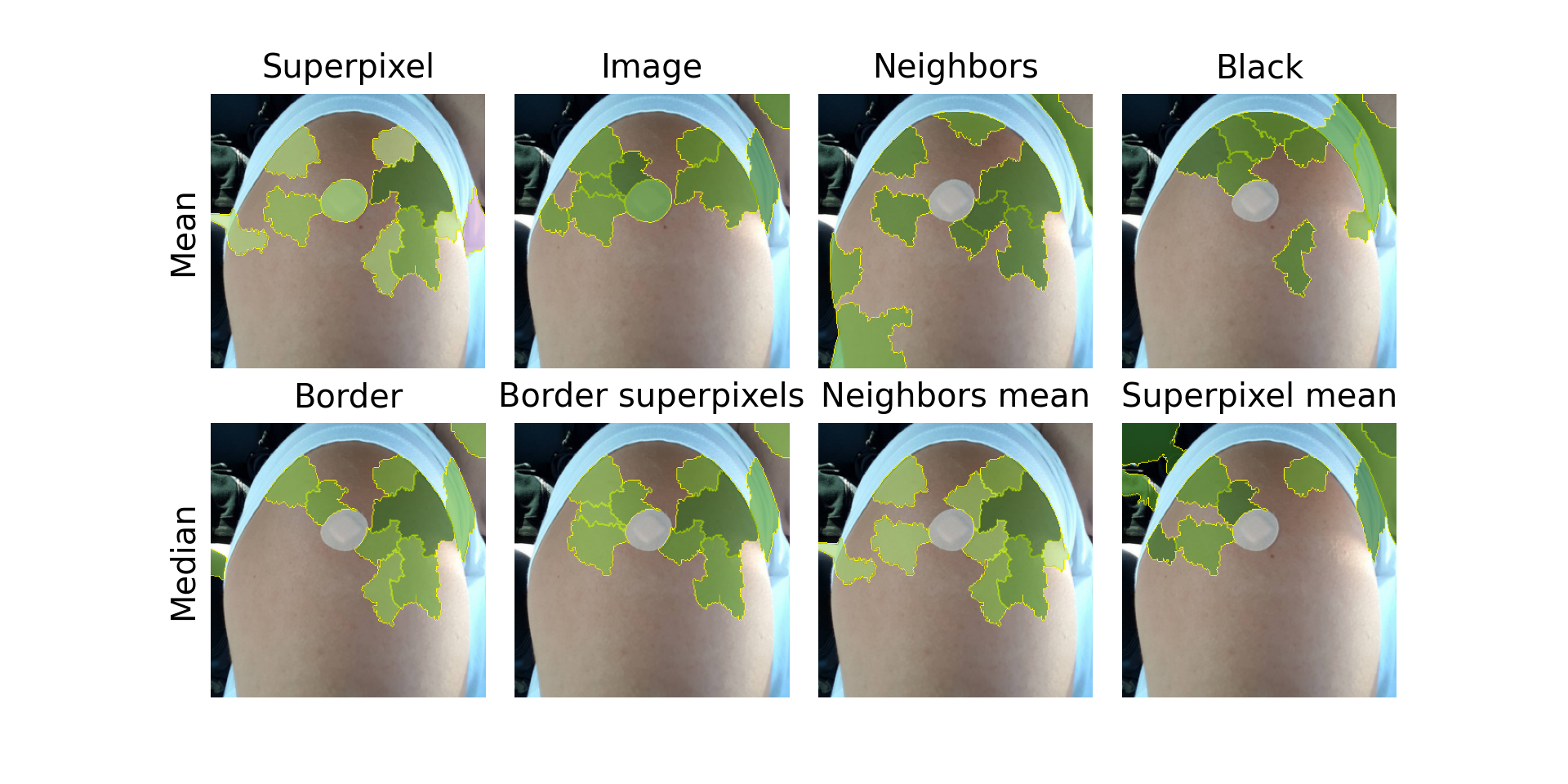}
\caption{
\textbf{Explanations using 8 deterministic occlusion methods:} 
Applied to an image of class \textit{band-aid} where default LIME fails to identify the most important superpixel(s). 
The top ten superpixels by absolute weight are visualized (darker green for a stronger positive contribution). 
From left to right, we can see the occlusion methods 1-8 as described in Section~\ref{sec:deterministic-occlusion}.
}
\label{fig:deterministic_replacements}
\end{figure}

While these methods showed improvements in specific cases, 
they still produce visually unnatural samples and fail to generalize effectively.
LIME struggles to identify the feature's importance,
when the replacement color closely resembles the original superpixel,
or if the shape of the superpixel reveals information about its content.  
We present a result by the example of the \textit{band-aid} class:
In Fig.~\ref{fig:deterministic_replacements} none of these strategies identifies the
band-aid as most important superpixel. LILI, however, does (see Fig.~\ref{fig:explanation-compare-introduction}).
Additional results can be found on GitHub\footnoteref{fn:github-chaves}.

\newpage

\subsection{Resolution-robust Large Mask Inpainting with Fourier Convolutions (LaMa)} \label{sec:lama}

LaMa, introduced by Suvorov et al. \cite{9707077}, is a feed-forward inpainting network designed to handle high-resolution images,
large missing areas, and complex geometric shapes by using fast Fourier convolutions (FFC) \cite{10.5555/3495724.3496100}.
FFC split the input channel-wise to a local and a global branch and use channel-wise 
fast Fourier transformations on the global branch for an image width receptive field.
The model architecture of LaMa is similar to a ResNet with 3 downsampling blocks,
followed by 6-18 FFC residual blocks, followed by 3 upsampling blocks.
LaMa is trained on the Places and the CelebA-HQ datasets using real images and synthetically generated masks,
and loss function, that is a weighted sum of a gradient penalty, an adversarial loss,
a discriminator-based perceptual loss, and a newly introduced loss which they call high receptive field perceptual loss.
Despite its relatively lightweight architecture (27 million parameters),
LaMa consistently outperforms older models like DeepFill v2 \cite{yu2018generative} in perceptual quality metrics such as
Fréchet Inception Distance (FID) \cite{NIPS2017_8a1d6947} and LPIPS \cite{8578166}. 

We chose LaMa for inpainting due to its balance between perceptual quality and computational efficiency,
making it well-suited for generating the large number of perturbed samples required by LIME (typically 1000 per explanation).
While state-of-the-art inpainting models based on transformers \cite{Elharrouss_2025} 
or diffusion architectures \cite{kim2026radregionawarediffusionmodels} achieve superior perceptual quality,
they are computationally expensive. For instance, even the region-aware diffusion model (RAD) \cite{kim2026radregionawarediffusionmodels},
which is faster than earlier diffusion-based methods, has significantly higher inference times compared to LaMa.

\subsection{Related Work} \label{sec:related-work}

Perturbation-based explanation methods are widely used in XAI for computer vision \cite{s25134166}.
A key challenge in these methods is generating perturbed samples that remain realistic and align with the original data distribution
\cite{aghakishiyeva2025photorealistic,10.1007/978-3-031-63787-2_12}. 
Recent advancements in generative inpainting have shown promising results in addressing this issue
by creating realistic perturbations that preserve contextual consistency. 

Tritscher et al.~\cite{10.1007/978-3-031-63787-2_12} apply generative inpainting to 
Shapley-value-based explanations for anomaly detection with tabular data.
By replacing traditional feature occlusion methods with a denoising diffusion probabilistic model (TabDDPM \cite{10.5555/3618408.3619133}),
they demonstrate stable explanation quality across various datasets and models. 

Regarding image data, Agarwal and Nguyen~\cite{agarwal2020explaining} and Aghakishiyeva et al. \cite{aghakishiyeva2025photorealistic}
explore the capabilities of generative inpainting for the generation of perturbed samples.
By introducing LIME-G, Agarwal and Nguyen~\cite{agarwal2020explaining} integrate generative inpainting into LIME.
LIME-G replaces per superpixel mean-color occlusion with DeepFill-v1/v2 \cite{yu2018generative,yu2018free},
to generate perturbed samples that are plausible and better aligned with the original data distribution.
Their findings showed that inpainting-based perturbations improve explanation accuracy and robustness compared to traditional occlusion methods.
Likewise, Aghakishiyeva et al. \cite{aghakishiyeva2025photorealistic} explore
photorealistic inpainting for perturbation-based explanations in ecological monitoring. 
Their method uses refined masks from the Segment Anything Model (SAM) \cite{Kirillov_2023_ICCV}
and Stable Diffusion \cite{rombach2021highresolution} to replace or remove objects or background in drone images.
However, the computational cost of this method using SAM and Stable Diffusion, limits it's scalability. 

Building on the work of Agarwal and Nguyen~\cite{agarwal2020explaining},
LILI leverages LaMa (a more recent inpainting model) and mask expansion
to generate more realistic perturbed samples with enhanced occlusion. 
Compared to Aghakishiyeva et al.\ \cite{aghakishiyeva2025photorealistic}
LILI balances computational cost and realism, as LaMa is less computationally expensive than Stable Diffusion.

\section{\methodeName} \label{sec:approach}
 
As discussed previously, LILI adopts LIME as given by the original implementation (see Section~\ref{sec:lime}) 
and only adjusts the second step of the explanation process: the creation of perturbed samples.
Two major changes are applied for LILI. Firstly, the generative replacement of superpixels
by the LaMa inpainting model, which is explained in Section~\ref{sec:perturbation-by-inpainting}.
Secondly, the introduction of a hyperparameter $m$ to enable the expansion of perturbation masks, 
which is discussed in Section~\ref{sec:perturbation-mask-expansion}.

\subsection{Perturbation by inpainting}\label{sec:perturbation-by-inpainting}

Regarding the generative replacement of superpixels for perturbed samples for LIME, we proceed as follows.
Let the image be of height $H$ and width $W$. Further let $M^{(i)} \in \{0,1\}^{H \times W}$
be the perturbation mask with $M^{(i)}_{k,{\ell}} = 1$ iff the superpixel corresponding to index $(k,\ell)$
is masked (\emph{i.e.}, switched of).
LIME, if superpixel $J_j$ is switched off, replaces all pixels of $J_j$ with the superpixel's mean color (default).
LIME-G and LILI however, generate the perturbed samples $x_i$ using the inpainting model $\mathcal{I}$,
the original image $\xi$ and the mask $M^{(i)}$:
\begin{equation}
    \label{eq:mask-inpainting}
    x_i = \mathcal{I}(\xi, M^{(i)}).
\end{equation}

For LIME-G $\mathcal{I}$ is given by DeepFill, for LILI by LaMa respectively.
In our experiments LIME-G is reimplementated using the PyTorch implementation of the DeepFill model \cite{Ang_2021}.

\subsection{Perturbation mask expansion}\label{sec:perturbation-mask-expansion}

In this section, we will first motivate the establishment of the hyperparameter~$m$, 
before we explain the technical details of the method realization.

\begin{figure}[h!]
\begin{center}
\includegraphics[trim=0cm 0cm 0cm 0cm, clip=true,width=1\textwidth]{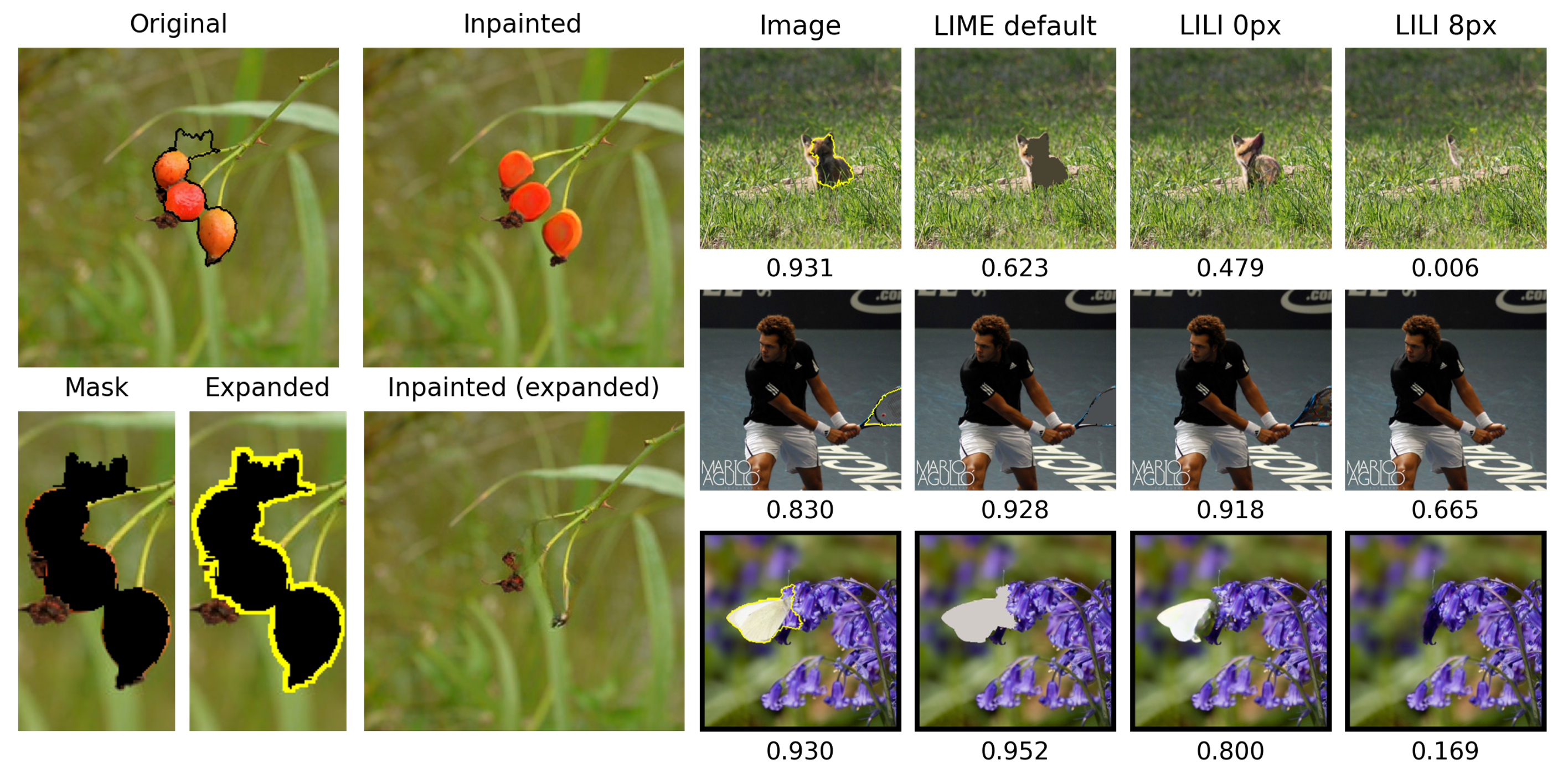}
\caption{
	\textbf{Motivation for perturbation mask expansion:} 
    The top-left image (\textit{original}) shows three hips and a perturbation mask (black outline). 
    The \textit{inpainted} sample closely resembles the original, but the \textit{mask} does not fully cover the hips.
    The \textit{expanded} mask hides remaining artifacts, encouraging background replacement (see \textit{inpainted expanded}).
    The right part shows perturbation effects on classification confidence (probabilities below).
    The \textit{image} column displays the original and masked region (\emph{i.e.}, yellow outline).
    \textit{LIME default} and \textit{LILI} with \textit{0 pixel} expansion leave the instance visible with high confidence,
    while \textit{LILI} with \textit{8 pixel} expansion removes it, causing a significant confidence drop.
}
\label{fig:mask-expansion}
\end{center}
\end{figure}

The introduction of the hyperparameter $m$ for mask expansion 
addresses limitations arising from the inpainting of superpixels
generated by LIME's default segmentation algorithm.
Segmentation methods, such as quickshift, tend to leave small artifacts at the border of objects \cite{pmlr-v151-garreau22a},
which can provide unintended cues to the inpainting model (see Fig.~\ref{fig:mask-expansion}).
For LaMa these artifacts can be sufficient to reconstruct the occluded object with high perceptual accuracy,
contradicting the intended removal of features for perturbation.
By expanding the perturbation mask by $m$ pixels around the original superpixel boundaries,
these artifacts can be hidden, forcing the inpainting model to generate content that aligns more closely with the background.
Fig.~\ref{fig:mask-expansion} illustrates this effect: The masked area here includes three \textit{rose hips}.
For effective feature occlusion we would expect them to be replaced (\emph{e.g.}, with a plausible background).
Without mask expansion, however, LaMa reconstructs the rose hips based on residual artifacts.
By using mask expansion, on the other hand, the rose hips are successfully removed.
The right part of the figure shows more examples of occlusion using the different methods. 
Looking at the classification confidence (annotated below) reveals that the occlusion 
using LILI with perturbation mask expansion, results in a significant drop.
Supported by these examples, we hypothesize that this approach enhances the effectiveness
of occlusion for perturbation-based sample generation.

Regarding the hyperparameter for perturbation masks expansion, we proceed as follows.
Instead of computing the mask expansion for each perturbation mask, we compute
the expanded masks $E^{(J_j)} \in \{0,1\}^{H \times W}$ for every superpixel $J_j$. 
This can be done prior to the sampling process and is 
computationally more efficient - as usually the number of perturbed samples is much larger than the number of superpixels.
The algorithm follows three major steps: 
With $m$ pixel mask expansion, we compute the expanded superpixel masks as follows. For $J_j \in J_1, \dots, J_d$:

\begin{samepage}
\begin{enumerate}
    \item Find the contour $C$ of $J_j$ using the border following algorithm of Suzuki and Abe \cite{SUZUKI198532}.
    \item Set $E_{k,\ell}^{(J_j)} = 1$ iff $(k,\ell) \in J_j$.
    \item Follow the contour for $(k,{\ell}) \in C$ and set $E_{r,s}^{(J_j)} = 1$ 
    for every pixel $(r,s)$ in the square of $m$ pixel around $(k,{\ell})$.
\end{enumerate}
\end{samepage}

The final expanded perturbation mask $\mathcal{M}^{(i)}_{k,{\ell}} \in \{0,1\}^{H\times W}$
for the $i$th perturbed sample is given by the element-wise disjunction (logical or)
of the expanded superpixel masks $E^{(J_j)}$ of all superpixels $J_j$ that are switched off (where $z_{i,j} = 1$).
\begin{figure}[h!]
\begin{center}
\includegraphics[clip=true,width=1\textwidth]{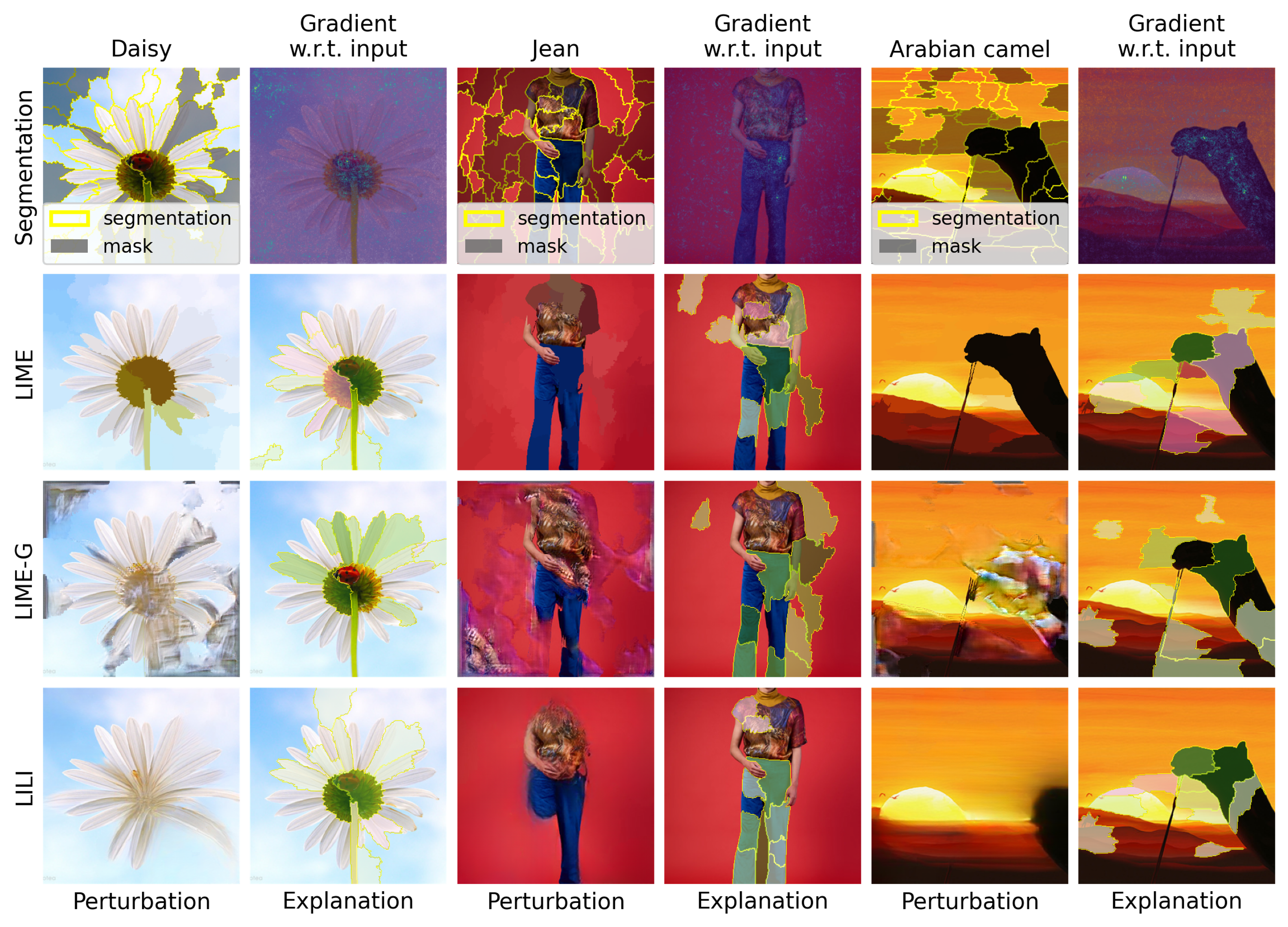}
\caption{
	\textbf{Examples of perturbed samples and explantions by methods:} 
    For each explained image, we can see the original with it's \textit{segmentation} and a perturbation mask on the left,
    and the gradients with respect to the input on the right (first row).
    Following rows show one perturbed sample, and the explanation as given by the corresponding methods.
}
\label{fig:perturbation-explanation-examples}
\end{center}
\end{figure}

\section{Evaluation} \label{sec:evaluation}

The proposed method, LILI, is evaluated alongside LIME and LIME-G across several criteria:  
inpainting quality (Section~\ref{sec:evaluation-inpainting-quality}),
explanation quality (Section~\ref{sec:explanation-quality}), 
explanation stability (Section~\ref{sec:evaluation-stability}), 
and computation time (Section~\ref{sec:explanation-times}). 
For the generation of explantions we generally use the default LIME hyperparameters, 
with adjustments for generating perturbed samples for LILI and LIME-G as described in Section~\ref{sec:approach}.
Images are selected randomly from the ImageNet 1k dataset of 
the ILSVRC 2012 challenge \cite{DBLP:journals/corr/SzegedyVISW15}.
For each image, we explain the class with the highest probability as
predicted by the InceptionV3 model \cite{DBLP:journals/corr/SzegedyVISW15}, trained on the ImageNet-1k dataset.

\subsection{Inpainting Quality} \label{sec:evaluation-inpainting-quality}
The LaMa model, used in LILI, consistently outperforms DeepFill (used in LIME-G) 
in perceptual inpainting quality across datasets and metrics \cite{9707077}.
To validate this for LIME perturbation mask inpainting, we conducted evaluations
using the Fréchet Inception Distance (FID)~\cite{NIPS2017_8a1d6947}, to quantify
the dissimilarity between real images and inpainted perturbed samples.

The FID compares the distributions of high level image features given by the activations of a classification CNN. 
As originally proposed \cite{NIPS2017_8a1d6947}, we compute the activations as given by the 2048-dimensional activation vectors
of the last pooling layer of the InceptionV3 network using the PyTorch FID package \cite{mseitzer_2024}.
The FID (see Eq.~\ref{eq:fid}) is then given by the Fréchet distance between the multidimensional Gaussian distributions
fitted to the activations from real images $\mathcal {N}(\mu ,\Sigma)$ and generated images ${\mathcal {N}}(\mu_g,\Sigma_g)$,
where $\mu$ and $\Sigma$ are the estimated mean vector and covariance matrix of the distribution of real images
($\mu_g$, $\Sigma_g$ of the generated distribution respectively):
\begin{equation}
    \label{eq:fid}
	d_{F} \left({\mathcal {N}}(\mu ,\Sigma),{\mathcal {N}}(\mu_g,\Sigma_g) \right)^{2} = 
	\lVert \mu -\mu_g\rVert _{2}^{2}
	+  \operatorname {trace} \left(\Sigma +\Sigma_g-2\left(\Sigma \Sigma_g\right)^{\frac {1}{2}}\right).
\end{equation}

We compute the FID from a reference set of 10,000 real images and compare it to a second set,
that consists of 5 perturbed samples for each of the real images (50,000 images total).
We repeat this for LILI, LIME-G and LIME perturbations.
Table~\ref{tab:fid} shows that LILI achieves the lowest FID score (6.727), 
significantly outperforming LIME-G (56.524) and LIME (30.532).
\begin{table}[h!]
    \centering
    \caption{
        \textbf{Fréchet inception distance (FID) of perturbed samples:} As generated by LIME, LILI and LIME-G.
        Lower values indicate a higher perceptual image quality.
    }
    \begin{tabular}{lrr}
                    & Occlusion method                    & FID \\
    \hline
    LIME   & superpixel mean color                         & 30.532                           \\ 
    LIME-G & inpainting with DeepFill                                      & 56.524                           \\
    LILI   & inpainting with LaMa                                          & \underline{6.727}                \\
    \end{tabular}
    \label{tab:fid}
\end{table}

\subsection{Explanation Quality}
\label{sec:explanation-quality}
As a proxy for explanation quality, we use the saliency metric \cite{10.5555/3295222.3295440}.
This metric measures how well attribution maps identify regions critical to the model's predictions
by computing the smallest rectangular patch containing the most salient regions and measuring
the classification probability for the cropped patch. A detailed description follows:

For thresholds $t = \alpha \mu_{max}$ given by $\alpha \in \{0, \ 0.05, \ 0.1, \ 0.15, \ \dots, \  0.95 \}$ and
the maximum of the superpixel weights $\mu_{max}$, the smallest rectangular image patch is cropped,
such that no saliency above the threshold is outside the patch. 
Now the ratio $a_t$ of the area of the patch over the area of the image is computed.
Afterwards, the patch is resized to the models input size (independently of the previous aspect ratio) 
and the classification probability $f(x_t)$ of the upsampled patch $x_t$ is computed.
Finally, the saliency metric is given as: 
\begin{equation}
    s(a_t, f(x_t)) = \log(\max(a_t, 0.05)) - \log(f(x_t)).
\end{equation}

Interpreted from an information theory perspective, this metric can be understood as the concentration of information
in the cropped image \cite{10.5555/3295222.3295440}. Lower saliency scores indicate better explanations.

The experiments were conducted with 100 randomly selected images from the ImageNet-1k test set,
using mask expansions of $0$, $3$, $5$, and $8$ pixels for both LILI and LIME-G.
For every method $\nu$ (LIME, LIME-G, LILI), mask expansion by $m \in \{0, 3, 5, 8\}$, 
image $i$ and threshold $\alpha$, the experiment yields a saliency score $S_{\nu,m,i,\alpha}$. 
We aggregate across thresholds to compute a score $S_{\nu,m,i}^{\star}$:
\begin{equation}
    \label{eq:saliency-min-selection}
    S_{\nu,m,i}^{\star} = \min_{\alpha} S_{\nu,m,i,\alpha}.
\end{equation}

Table~\ref{tab:saliency-metric-min-value-selection} shows that LILI with a 3-pixel mask expansion
achieves the best saliency scores across most statistical measures, outperforming both LIME and LIME-G.
This demonstrates that LILI provides faithful and robust explanations.
Agarwal and Nguyen \cite{agarwal2020explaining} showed that LIME-G performs on-par with LIME regarding the saliency metric.
Our improved results therefore are of particular interest.
\begin{table}[h!]
\centering
\caption{
    \textbf{Statistics of the saliency metric scores:} 
    For each method and mask expansion is presented: 
    mean, first quartile (Q1), median, third quartile (Q3), interquartile range (IQR) and median absolute deviation (MAD).
    Lower values indicate better explanations.
    The best result (minimum aggregated saliency score) for mean, Q1, median and Q3 is underlined.
}
\label{tab:saliency-metric-min-value-selection}
\begin{tabular}{llrrrrrr}
 &  & mean & Q1 & median & Q3 & IQR & MAD \\
Method & Mask expansion &  &  &  &  &  &  \\
\hline
LILI & 0 & -1.766 & -2.410 & -1.981 & -1.159 & 1.251 & 0.574 \\
 & 3 & \underline{-1.816} & -2.427 & \underline{-2.011} & \underline{-1.357} & \underline{1.070} & \underline{0.544} \\
 & 5 & -1.733 & \underline{-2.451} & -1.917 & -1.245 & 1.206 & 0.619 \\
 & 8 & -1.692 & -2.427 & -1.875 & -1.189 & 1.238 & 0.632 \\
\hline
LIME-G & 0 & -1.637 & -2.387 & -1.774 & -1.013 & 1.374 & 0.691 \\
 & 3 & -1.699 & -2.427 & -1.864 & -1.098 & 1.329 & 0.701 \\
 & 5 & -1.647 & -2.349 & -1.737 & -1.059 & 1.290 & 0.636 \\
 & 8 & -1.744 & -2.367 & -1.944 & -1.187 & 1.180 & 0.630 \\
\hline
LIME & 0 & -1.625 & -2.308 & -1.756 & -1.188 & 1.119 & 0.559 \\
\end{tabular}
\end{table}

\subsection{Explanation Stability}
\label{sec:evaluation-stability}
\begin{figure}[h!] 
\begin{center}
\includegraphics[clip=true,width=0.66\textwidth]{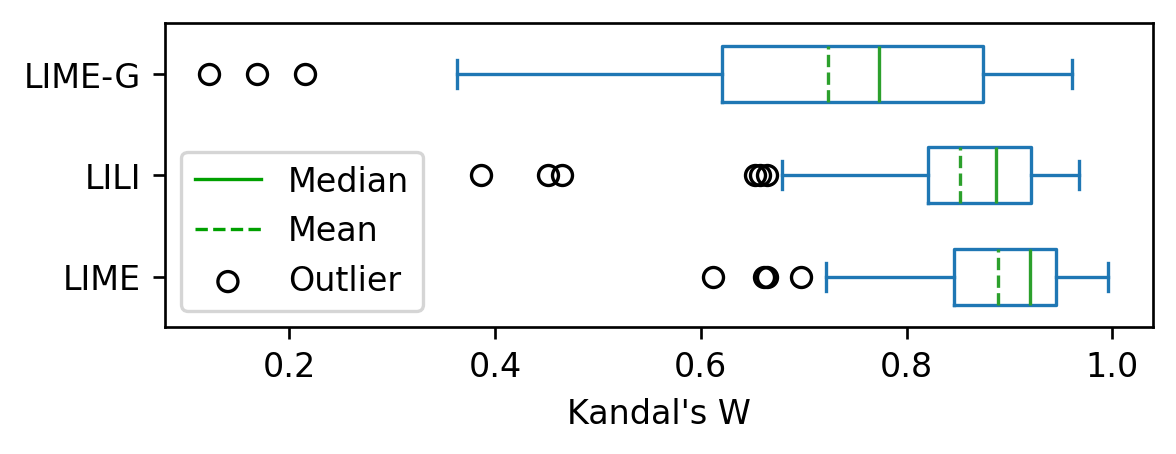}
\caption[Kendall's W of the top 10 superpixels]{
    \textbf{Boxplot of the agreement of rankings:} 
    As computed using Kendall's W (coefficient of concordance) 
    for the top 10 superpixels (by absolute mean weight) for each of the 100 sample images.
}
\label{fig:explanation-stability-kandal}
\end{center}
\end{figure}

The stability of explanations was assessed by computing 10 explanations for each of 100 images,
using the same superpixel segmentation but varying perturbed samples.
Note that for comparability for both LIME-G and LILI, the perturbation mask is expanded by 5 pixels.
To summarize across all images, the coefficient of concordance (Kendall's W \cite{10.1214/aoms/1177732186}) 
is computed for the top 10 superpixels with highest absolute mean value.
This can be interpreted as a measure of the agreement of rankings across different runs, 
with 0 for no agreement to 1 for complete agreement.
As shown in Table~\ref{tab:explanation-stability-kandal}, 
LIME achieves the highest mean concordance (0.889), followed by LILI (0.852) and LIME-G (0.723).
The boxplot in Figure~\ref{fig:explanation-stability-kandal} demonstrates that LILI, 
with the exception of three outliers, achieves significantly higher explanation stability than LIME-G,
while LIME exhibits only a marginal improvement over LILI.
\begin{table}[h!]
\centering
\caption[Kendall's W of top 10 superpixels]{
    \textbf{Agreement of rankings:}
    As computed using Kendall's W (coefficient of concordance) 
    for the top 10 superpixels (by absolute mean weight) for each of the 100 sample images.
    The table shows the aggregated results as mean and standard deviation (std) across all samples.
}
\label{tab:explanation-stability-kandal}
\begin{tabular}{lrrr}
 & LIME & LILI & LIME-G \\
\hline
mean & 0.889 & 0.852 & 0.723 \\
std & 0.080 & 0.109 & 0.184 \\
\end{tabular}
\end{table}

\subsection{Computation Time}
\label{sec:explanation-times}
The execution times for generating explanations were tracked for 100 images using a NVIDIA L40 GPU for model inference.
Table~\ref{tab:explanation-execution-times} shows that LIME is the fastest method,
followed by LIME-G and LILI.
The increased runtimes for LILI and LIME-G are attributed to the computational cost of inpainting.
\begin{table}[h!]
\centering
\caption[Execution times of explanations]{
    \textbf{Execution times of explanations:}
    The table shows the mean and the standard deviation (std) of the execution times in seconds of the generation of explanations.
}
\label{tab:explanation-execution-times}
\begin{tabular}{llrr}
 &  & mean & std \\
Method & Mask expansion &  &  \\
\hline
LIME & 0 & 4.431 & 0.099 \\
\hline
LIME-G & 0 & 8.708 & 0.037 \\
 & 3 & 8.915 & 0.044 \\
 & 5 & 9.008 & 0.046 \\
 & 8 & 9.154 & 0.055 \\
\hline
LILI & 0 & 12.126 & 0.214 \\
 & 3 & 12.334 & 0.092 \\
 & 5 & 12.425 & 0.062 \\
 & 8 & 12.575 & 0.073 \\
\end{tabular}
\end{table}

\section{Discussion} \label{sec:discussion}
While the current experimental setup utilizing LIME's default hyperparameters, the InceptionV3 model,
and the saliency metric for ImageNet provides a solid and reproducible baseline, 
it also opens promising avenues for further exploration.
This could be extended by analyzing alternative models, diverse datasets,
and more nuanced evaluation metrics to capture a broader spectrum of explanation quality.
Since LIME is known to be sensitive to hyperparameter changes \cite{DBLP:journals/corr/abs-2010-12487}, 
future work will in particular include the evaluation of the generalizability of the results
for different hyperparameter combinations.

Our results show that LIME-G and LILI produce less stable explanations than LIME, which we hypothesize stems from two main factors:
Firstly, for LIME, one masked superpixel is always replaced by one deterministic color (\emph{e.g.}, superpixel mean). 
LILI, however, inpaints superpixels depending on the perturbation mask, which is determined by the combination of masked superpixels.
While this can be an advantage, if, \emph{e.g.}, an object consists of two separate superpixels being only erased if both are inpainted,
it might influence the explanation stability.
Secondly, mask expansion introduces overlapping superpixels, which can influence neighboring regions, causing explanations to be less coherent.
This highlights the need for careful tuning of the mask expansion parameter to balance effective occlusion and unintended interactions.

\section{Conclusion} \label{sec:conclusion}
In this work, we introduced \methodeName (LILI), a novel approach to improve perturbation-based explanations
by generating more realistic perturbed samples using the LaMa inpainting model. 
The Fréchet Inception Distance (FID) results confirm that LILI, compared to LIME and LIME-G,
produces perturbed samples with higher perceived image quality,
that are better aligned with the original data distribution, fostering trust by reducing the risk for unexpected model behaviour. 
LILI achieves, compared to LIME and LIME-G, superior explanation quality, as demonstrated by the saliency metric,
particularly when combined with the proposed hyperparameter for mask expansion.
However, while LILI offers improvements in explanation quality, it does so with computational costs that are higher than LIME
but remain lower than \emph{e.g.}, diffusion-based approaches, such as Aghakishiyeva et al.~\cite{aghakishiyeva2025photorealistic}.
This balance between quality and efficiency makes LILI an alternative for perturbation-based explanations in scenarios
where computational resources are limited.

Future work will include additional evaluation metrics and models to rigorously validate explanation quality.
We will explore the impact of hyperparameters and integrate state-of-the-art inpainting models.
Moreover, extending LILI to other modalities and perturbation-based explanation frameworks
will provide further insights into its generalizability and applicability across diverse machine learning tasks.

\subsubsection{Acknowledgements} 
This work has been supported by the French government through the NIM-ML project (ANR-21-CE23-0005-01).
All experiments were performed using the Julia 2 cluster,
funded as DFG project as ``Forschungsgro{\ss}ger{\"a}t nach Art 91b GG'' under INST 93/1145-1 FUGG.

\bibliographystyle{splncs04}
\bibliography{bib}

\end{document}